%% file: main.tex
\documentclass[letterpaper, 10 pt, conference]{ieeeconf}  

\IEEEoverridecommandlockouts                              

\usepackage{times} 
\usepackage{amsmath} 
\usepackage{amssymb}  
\usepackage{amsfonts}
\usepackage[dvipsnames]{xcolor}
\usepackage[utf8]{inputenc}
\usepackage{graphicx}
\usepackage[hyphens]{url}
\usepackage{soul}
\usepackage{booktabs}
\usepackage[ruled,vlined, linesnumbered]{algorithm2e}
\usepackage{multirow}
\usepackage[T1]{fontenc}
\usepackage{balance}
\usepackage{float}
\usepackage{tcolorbox}
\tcbuselibrary{listings,skins,breakable,theorems}
\let\labelindent\relax
\usepackage{enumitem}
\usepackage{mathtools}
\usepackage{hyperref}

\input{macros}

\title{\LARGE \bf
Optimization-Based Robust Permissive Synthesis for Interval MDPs
}

\author{
    Khang Vo Huynh\thanks{Khang Vo Huynh and Lu Feng are with the Department of Computer Science, University of Virginia, Charlottesville, VA, USA. Emails: \{szq2sj, lu.feng\}@virginia.edu}
    \and
    David Parker\thanks{David Parker is with the Department of Computer Science, University of Oxford, Oxford, UK. Email: david.parker@cs.ox.ac.uk}
    \and
    Lu Feng
}

\begin{document}

\maketitle
\thispagestyle{empty}
\pagestyle{empty}

\begin{abstract}
\input{0_abstract}
\end{abstract}

\section{Introduction} \label{sec:intro} 
\input{1_intro}

\subsection{Related Work} \label{sec:related} 
\input{2_related}

\section{Problem Formulation} \label{sec:problem} 
\input{3_problem}

\section{Methods} \label{sec:methods} 
\input{4_methods}

\section{Experiments} \label{sec:exp} 

\input{5_exp}

\section{Conclusion} \label{sec:conclu} 
\input{6_conclu}

\section*{Appendix} \label{sec:appendix} 
\input{7_appendix}


\bibliographystyle{IEEEtran}
\bibliography{references}

\end{document}

%% file: macros.tex
\newcommand{\sectref}[1]{Section~\ref{#1}}

\newcommand{\figref}[1]{Figure~\ref{#1}}
\newcommand{\tabref}[1]{Table~\ref{#1}}
\newcommand{\egref}[1]{Example~\ref{#1}}
\newcommand{\encref}[1]{Encoding~\ref{#1}}

\newcommand{\thmref}[1]{Theorem~\ref{#1}}

\newcommand{\lemref}[1]{Lemma~\ref{#1}}

\newtheorem{lemma}{Lemma}
\newtheorem{theorem}{Theorem}

\newcommand{\startpara}[1]{{\vskip5pt\noindent{\bf #1.}}} 
\renewcommand{\url}[1]{{\def~{\char126}\sf#1}}

\newcounter{exampcount}
\newenvironment{examp}
{\refstepcounter{exampcount}
\vskip6pt\noindent
{\textit{Example \arabic{exampcount}}.}}
{\hfill\scalebox{0.85}{$\blacksquare$}\vskip4pt}

\newtcolorbox{problemBox}[2][]{colback=red!5!white,
                               colframe=red!50!black,
                               boxrule=0.8pt,
                               arc=2pt,
                               left=1pt,
                               right=1pt,
                               top=1pt,
                               bottom=1pt,
                               enhanced,
                               title={Problem: #2}}

\newenvironment{problem}[1]
  {\begin{problemBox}{#1}}
  {\end{problemBox}}

\newcounter{encoding} 
\renewcommand{\theencoding}{\arabic{encoding}}

\newtcolorbox{encodingBox}[2][]{colback=blue!5!white,
                                colframe=blue!50!black,
                                boxrule=0.8pt,
                                arc=2pt,
                                left=1pt,
                                right=1pt,
                                top=1pt,
                                bottom=1pt,
                                enhanced,
                                title={Encoding~\theencoding: #2},
                                float=tp,   
                                }

\newenvironment{encoding}[1]
  {\refstepcounter{encoding}\begin{encodingBox}{#1}}
  {\end{encodingBox}}

\DeclareMathOperator*{\maximize}{maximize}



%% file: 0_abstract.tex
We present an optimization-based framework for robust permissive synthesis for Interval Markov Decision Processes (IMDPs). While robust IMDP controller synthesis typically yields a single policy and most permissive-synthesis methods assume exact transition models, we synthesize multi-strategies that retain multiple actions while guaranteeing satisfaction of probabilistic reachability or expected-reward specifications under all admissible transition probabilities. We formulate the problem as a mixed-integer linear program (MILP) that maximizes the number of enabled state--action pairs subject to robust Bellman constraints. We develop two encodings: a direct vertex-enumeration formulation and a dualization-based formulation that avoids explicit enumeration of uncertainty-polytope vertices and has size linear in the number of successor transitions. Experiments on four benchmark domains show that both encodings achieve the same optimal permissiveness and scale to IMDPs with hundreds of thousands of states. Compared with standard robust single-policy synthesis, the resulting multi-strategies retain substantially more action choices.

%% file: 1_intro.tex
Decision-making under transition uncertainty arises in many control and planning problems. Standard Markov decision processes (MDPs) assume precise transition probabilities, whereas Interval MDPs (IMDPs) model epistemic uncertainty by allowing transition probabilities to vary within specified intervals~\cite{badings2023decision,suilen2024robust}. Robust controller synthesis for IMDPs has therefore been studied extensively for reachability, temporal-logic, and reward objectives~\cite{nilim2005robust,iyengar2005robust,wolff2012robust}. These methods, however, typically synthesize a single deterministic or randomized policy.

Permissive synthesis instead returns a \emph{multi-strategy} that enables multiple actions at each state while guaranteeing that every compliant strategy satisfies the specification. Such controllers preserve flexibility for subsequent decision making and can, for example, serve as safety filters for learning-based controllers~\cite{drager2015permissive,jansen2020safe}. Existing permissive-synthesis methods largely assume precise transition probabilities, although recent shielding work has considered IMDPs under transition uncertainty~\cite{reed2025learning}.

In this paper, we study robust permissive synthesis for IMDPs under probabilistic reachability and expected-reward specifications. Given an IMDP, we seek a multi-strategy that maximizes the number of enabled state--action pairs while guaranteeing specification satisfaction for every compliant strategy and every admissible transition realization. We formulate this problem as a mixed-integer linear program (MILP) and develop two alternative encodings:
\begin{enumerate}
    \item a \emph{vertex-enumeration encoding} that directly enforces the robust Bellman constraints over the vertices of each transition-uncertainty polytope; and
    \item a \emph{dualization-based encoding} that eliminates explicit vertex enumeration using linear-programming duality, yielding an encoding whose size grows linearly with the number of successor transitions.
\end{enumerate}

We implement both encodings and evaluate them on four IMDP benchmark domains. The two encodings achieve the same optimal permissiveness but exhibit different computational tradeoffs depending on model structure. We additionally compare against standard robust single-policy synthesis and show that permissive synthesis retains substantially more action choices across all benchmark instances.

%% file: 2_related.tex
\startpara{Robust synthesis for IMDPs}
IMDPs have been widely used for planning and control under transition uncertainty, including robotic navigation, stochastic systems, and learned abstractions~\cite{jiang2022safe,jiang2023abstraction,cauchi2019efficiency,adams2022formal,mathiesen2025scalable}. Robust synthesis for IMDPs has been studied extensively for reachability, temporal-logic, and multi-objective specifications~\cite{nilim2005robust,iyengar2005robust,wolff2012robust,puggelli2013polynomial,hahn2017multi}. These methods typically synthesize a single deterministic or randomized policy.

\startpara{Permissive synthesis}
Permissive synthesis retains multiple actions while guaranteeing specification satisfaction. Existing approaches include MILP-based synthesis for MDPs~\cite{drager2015permissive}, multi-objective permissive control~\cite{chen2022multi}, and related shielding methods for safe reinforcement learning~\cite{junges2016safety,jansen2020safe,alshiekh2018safe}. Most assume precise transition models.
Recent work by Reed and Lahijanian~\cite{reed2025learning} also uses IMDPs to retain safe actions under transition uncertainty for shielding unknown continuous systems. Their notion of maximality is set-based: the shield retains a maximal set of policies satisfying a safety threshold. Our formulation instead optimizes a quantitative permissiveness objective, maximizing the total number of enabled state--action pairs over satisfying multi-strategies.

%% file: 3_problem.tex
\startpara{Interval MDPs (IMDPs)}
Formally, an IMDP is a tuple 
$\mathcal{M} = (S, s_{\iota}, A, \check{P}, \hat{P}, R)$,
where $S$ is a finite set of states with initial state $s_{\iota} \in S$,
$A$ is a finite set of actions, 
$\check{P}, \hat{P}: S \times A \times S \to [0,1]$ assign lower and upper bounds
to each transition, and $R: S \times A \to \mathbb{R}_{\ge 0}$ is a reward function.  
Intuitively, an IMDP represents a family of MDPs. 
For each state-action pair $(s,a)$, the admissible successor distributions are
\(
\mathcal{P}(s,a) = \{\, P \in \Delta(S) \mid 
\check{P}(s,a,s') \le P(s,a,s') \le \hat{P}(s,a,s') \;\; \forall s' \in S \,\},
\)
where $\Delta(S)$ is the set of all discrete probability distributions over $S$.
We define $\mathcal{P}$ as the set of all transition functions 
$P : S \times A \to \Delta(S)$ with $P(s,a,\cdot) \in \mathcal{P}(s,a)$ 
for every $(s,a)$.

\startpara{Permissive Controller}
An IMDP is controlled by a \emph{strategy} (also called a \emph{policy}), which in this 
work is assumed to be deterministic and memoryless. 
Formally, a strategy is a function $\sigma : S \to A$ such that 
$\sigma(s) \in \alpha(s)$, where $\alpha(s) \subseteq A$ denotes the set of 
\emph{enabled actions} in state $s$. 
A \emph{permissive controller}, or \emph{multi-strategy}, generalizes this notion 
by allowing a set of actions at each state; formally, it is a mapping 
$\theta : S \to 2^A$ such that $\theta(s) \subseteq \alpha(s)$ and 
$\theta(s) \neq \emptyset$ for all $s \in S$. 
For a multi-strategy $\theta$, define its \emph{permissiveness} as
$\beta(\theta) \coloneqq \sum_{s\in S}\sum_{a\in \alpha(s)} \mathbf{1}[a \in \theta(s)]$.
This definition assigns equal weight to all enabled state--action pairs. 
When application-specific preferences are available, it can be generalized to 
a weighted permissiveness measure
$\sum_{s\in S}\sum_{a\in\alpha(s)} w_{s,a}\mathbf{1}[a\in\theta(s)]$ 
with $w_{s,a}\ge 0$.
A strategy $\sigma$ is said to be \emph{compliant} with a multi-strategy $\theta$, 
denoted $\sigma \triangleleft \theta$, 
if $\sigma(s) \in \theta(s)$ for every $s \in S$.

\startpara{Specifications}
We consider two common types of properties.  
A \emph{probabilistic reachability} property has the form 
$\mathsf{P}_{\bowtie p}(\Diamond T)$, where $T \subseteq S$ is a target set, 
$\Diamond$ denotes eventual reachability, $p \in [0,1]$, and 
$\bowtie \in \{\le,\ge\}$; it requires that the probability of eventually 
reaching $T$ satisfies the bound $\bowtie p$.  
An \emph{expected total reward} property has the form 
$\mathsf{R}_{\bowtie b}(\Diamond T)$, where $b \in \mathbb{R}_{\ge 0}$, meaning 
the expected total reward accumulated until reaching $T$ satisfies the bound 
$\bowtie b$.  
More complex temporal specifications can often be reduced to reachability 
objectives through standard product constructions.

\startpara{Satisfaction Semantics}
Given an IMDP $\mathcal{M}$, a property $\varphi$, a strategy $\sigma$, and a 
transition function $P \in \mathcal{P}$, we write 
$(\mathcal{M},\sigma,P) \models \varphi$ if $\varphi$ holds in the 
MDP obtained by resolving all choices in $\mathcal{M}$ using 
strategy $\sigma$ and transition function $P$. For a multi-strategy $\theta$, 
$(\mathcal{M},\theta,P) \models \varphi$ if this holds for all 
$\sigma \triangleleft \theta$. Robust satisfaction then quantifies over all 
admissible transitions: $(\mathcal{M},\sigma) \models_{\text{rob}} \varphi$ if 
$(\mathcal{M},\sigma,P) \models \varphi$ for all $P \in \mathcal{P}$, and 
$(\mathcal{M},\theta) \models_{\text{rob}} \varphi$ if 
$(\mathcal{M},\theta,P) \models \varphi$ for all $P \in \mathcal{P}$.

\startpara{Problem Statement}
We aim to synthesize a \emph{robust permissive controller} 
that guarantees satisfaction of the specification under all admissible 
uncertainties while maximizing the number of enabled action choices. 

\begin{problem}{Robust Permissive Controller Synthesis}
Given an IMDP $\mathcal{M} = (S, s_{\iota}, A, \check{P}, \hat{P}, R)$ and a 
specification $\varphi$ of the form 
$\mathsf{P}_{\bowtie p}(\Diamond T)$ or $\mathsf{R}_{\bowtie b}(\Diamond T)$, 
synthesize a multi-strategy $\theta$ such that:
\begin{enumerate}[label=\arabic*.]
    \item (\textbf{Robust satisfaction}) 
    $(\mathcal{M},\theta) \models_{\text{rob}} \varphi$.
    \item (\textbf{Optimal permissiveness}) 
    $\beta(\theta) \geq \beta(\theta')$ for every multi-strategy $\theta'$ 
    satisfying $(\mathcal{M},\theta') \models_{\text{rob}} \varphi$.
\end{enumerate}
\end{problem}

\begin{figure}
    \centering
    \includegraphics[width=0.7\linewidth]{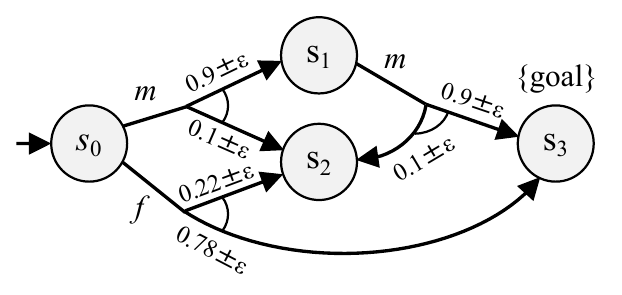}
    \caption{A simple robotic navigation task modeled as an IMDP.}
    \label{fig:example}
\end{figure}

\begin{examp} \label{eg:imdp}
Consider a robotic navigation task modeled as the IMDP $\mathcal{M}$ in \figref{fig:example}.
A robot starts in $s_0$ and must reach the goal state $s_3$. From $s_0$, it may 
choose a \emph{fast} action ($f$), which can reach $s_3$ directly but with risk 
of failure into $s_2$, or a \emph{medium} action ($m$), which is safer but must 
be applied twice through $s_1$. 
Transition probabilities are given as intervals of radius $\varepsilon \in [0,1]$, 
a constant capturing the level of epistemic uncertainty. 
For example, $P(s_0,f,s_3) \in [0.78-\varepsilon,\,0.78+\varepsilon]$, clipped to $[0,1]$.  
A property such as $\varphi = \mathsf{P}_{\ge 0.65}(\Diamond s_3)$ asks whether the 
probability of eventually reaching the goal is at least $0.65$. 
The synthesis problem asks for an optimally permissive multi-strategy 
that robustly satisfies $\varphi$.  
\end{examp}

%% file: 4_methods.tex
We now present methods for solving the IMDP robust permissive controller 
synthesis problem. A straightforward approach is to 
explicitly enumerate all vertices of the uncertainty polytopes in the IMDP and
adopt existing permissive controller synthesis techniques for MDPs~\cite{drager2015permissive}.
This yields a mixed-integer linear 
program (MILP) encoding, but the number of vertices grows exponentially with 
the number of successor states per state–action pair. To address this scalability 
issue, we develop a dualization-based encoding that captures the adversarial 
choice of transition probabilities without explicit enumeration, yielding a 
compact MILP. We present the vertex enumeration encoding in \sectref{sec:vertex} 
and the dualization-based encoding in \sectref{sec:dual}.

\subsection{Vertex Enumeration Encoding} \label{sec:vertex}

We present an MILP for the probabilistic reachability property $\mathsf{P}_{\ge p}(\Diamond T)$, given in \encref{enc:vertex}.
The formulation introduces real-valued variables $x_s \in [0,1]$, representing lower bounds on the worst-case probability of reaching $T$ from state $s$, and binary variables $y_{s,a}$, indicating whether the multi-strategy $\theta$ admits action $a \in \alpha(s)$.

The objective function~(\ref{enc:obj-vertex}) maximizes the number of admitted actions, thereby ensuring optimal permissiveness.
Because the objective maximizes the total number of enabled state--action pairs, multiple distinct multi-strategies may attain the same optimal permissiveness.

Constraint~(\ref{enc:c1-vertex}) ensures that all states with enabled actions admit at least one of them.
Constraint~(\ref{enc:c2-vertex}) enforces that the initial state achieves the threshold $p$.
Constraint~(\ref{enc:c3-vertex}) fixes the value of all target states to~1.

Constraint~(\ref{enc:c4-vertex}) enforces robustness by requiring $x_s$ to hold for all feasible successor distributions of each non-target pair $(s,a)$. The term $M(1-y_{s,a})$ deactivates the constraint when action $a$ is not allowed, 
where $M$ is a sufficiently large constant (the standard big-$M$ method) used 
for linearization and constraint deactivation~\cite{mccormick1976computability}.
The vertex set $\mathcal{V}(s,a)$ consists of the extreme points of $\mathcal{P}(s,a)$, obtained by pushing transition probabilities to their interval bounds while maintaining that they remain nonnegative and sum to one. Each $v \in \mathcal{V}(s,a)$ defines a concrete distribution $P^{(v)}(s,a,\cdot)$, and enforcing the inequality for all $v$ ensures that $x_s$ is robust.

\input{4a_vertex}

\begin{figure}
    \centering
    \includegraphics[width=0.9\linewidth]{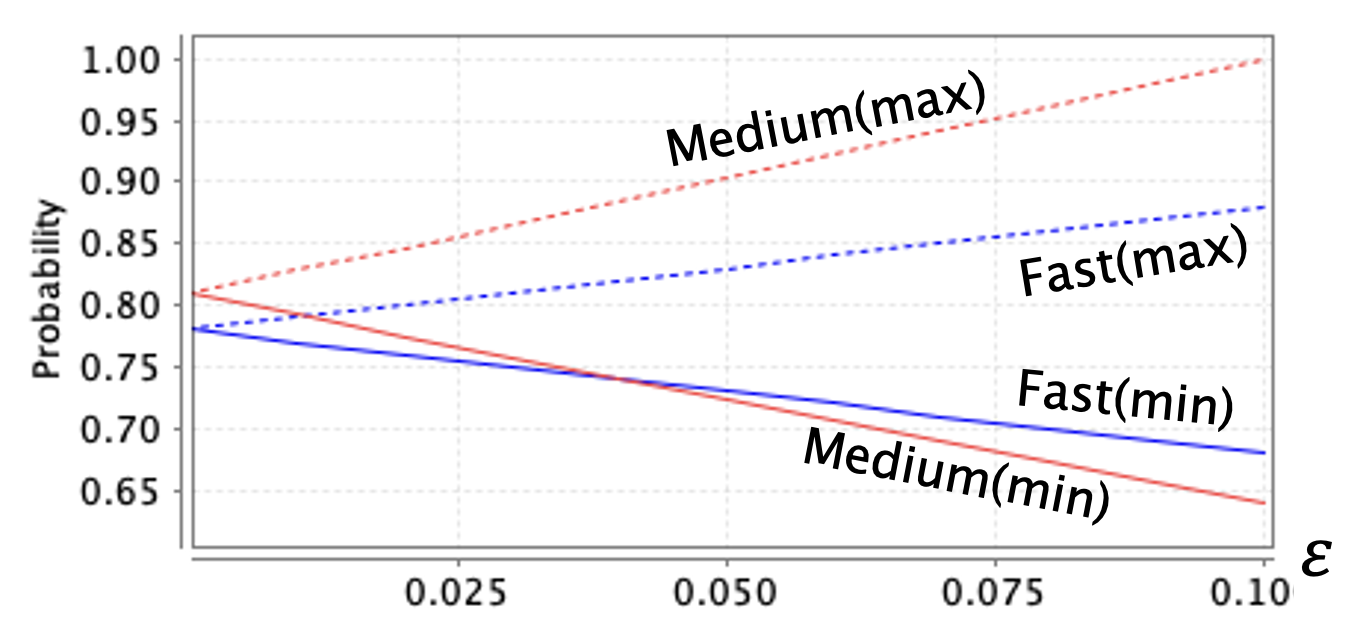}
    \caption{Robust reachability probability as the uncertainty radius $\varepsilon$ varies.}
    \label{fig:curves}
\end{figure}

\begin{examp} \label{eg:vertex}
We apply \encref{enc:vertex} to the IMDP in \figref{fig:example} 
with respect to the property $\varphi = \mathsf{P}_{\ge 0.65}(\Diamond s_3)$. 
Variables $x_s$ represent lower bounds on the worst-case reachability 
probability from each state,
while binary variables $y_{s_0,f}$, $y_{s_0,m}$, and $y_{s_1,m}$ specify 
which actions (fast or medium) are allowed by the multi-strategy in states $s_0$ and $s_1$. 
The objective is to maximize the sum $y_{s_0,f} + y_{s_0,m} + y_{s_1,m}$. 
Constraints~(\ref{enc:c1-vertex})–(\ref{enc:c3-vertex}) enforce that at least 
one action is chosen in $s_0$ and $s_1$, that $x_{s_0} \ge 0.65$, and that 
$x_{s_3}=1$.  
Constraint~(\ref{enc:c4-vertex}) is instantiated for each vertex of the 
uncertainty polytope. For example, for the fast action at $s_0$, 
$\mathcal{P}(s_0,f)$ reduces to an interval over successors 
$\{s_2,s_3\}$ and yields two inequalities:
\begin{align*}
x_{s_0} &\le (0.22+\varepsilon)\,x_{s_2} + (0.78-\varepsilon)\,x_{s_3} 
          + M(1-y_{s_0,f}), \\
x_{s_0} &\le (0.22-\varepsilon)\,x_{s_2} + (0.78+\varepsilon)\,x_{s_3} 
          + M(1-y_{s_0,f}).
\end{align*}

Solving the MILP shows how the resulting multi-strategy depends on the 
uncertainty radius~$\varepsilon$ and the threshold~$p$. For $\varepsilon=0.1$, 
requiring $\mathsf{P}_{\ge 0.65}(\Diamond s_3)$ admits only the fast action. 
Lowering the threshold to $p=0.6$ or reducing uncertainty to $\varepsilon=0.05$ 
admits both fast and medium actions. These cases are illustrated in \figref{fig:curves}, 
which plots the minimal (solid) and maximal (dotted) reachability probabilities 
for each action as the uncertainty radius $\varepsilon$ varies.
\end{examp}

\startpara{Correctness}
The correctness of the MILP in \encref{enc:vertex} is stated in the following 
theorem, with the proof provided in the appendix.

\begin{theorem} \label{thm:vertex}
Let $\mathcal{M}$ be an IMDP and 
$\varphi=\mathsf{P}_{\ge p}(\Diamond T)$ a property.
An optimal solution to the MILP in \encref{enc:vertex} induces a robust,
optimally permissive multi-strategy
$\theta(s)=\{a\in\alpha(s)\mid y_{s,a}=1\}$ for $\varphi$.
Conversely, every robust, optimally permissive multi-strategy can be encoded
by an optimal solution whose objective value equals
$\beta(\theta)=\sum_{s\in S}\sum_{a\in\alpha(s)}y_{s,a}$.
\end{theorem}

\startpara{Adapting \encref{enc:vertex} for Other Specifications}
For properties of the form 
$\mathsf{P}_{\le p}(\Diamond T)$, constraint~(\ref{enc:c2-vertex}) is 
replaced with $x_{s_\iota} \le p$, and 
constraint~(\ref{enc:c4-vertex}) becomes
\[
x_s \ge \sum_{s'\in S} P^{(v)}(s,a,s') \cdot x_{s'} - M(1-y_{s,a}).
\]
For expected total reward properties $\mathsf{R}_{\ge b}(\Diamond T)$, 
we instead use $x_s \in \mathbb{R}_{\ge 0}$ to denote reward-to-go 
variables, modify constraint~(\ref{enc:c2-vertex}) to 
$x_{s_\iota} \ge b$, set constraint~(\ref{enc:c3-vertex}) to $x_s=0$, 
and replace constraint~(\ref{enc:c4-vertex}) with
\[
x_s \le \sum_{s'\in S} P^{(v)}(s,a,s') \cdot x_{s'} 
+ r(s,a) + M(1-y_{s,a}),
\]
where $r(s,a)$ is the immediate reward of action $a$ in state $s$. 
Analogous changes apply to properties of the form 
$\mathsf{R}_{\le b}(\Diamond T)$ by reversing the inequalities accordingly.
For total-reward properties, we additionally require that $T$ is reached with probability~1, ensuring that the accumulated reward until reaching $T$ is well defined.

\startpara{Scalability}
Although conceptually straightforward, the vertex enumeration encoding 
can scale poorly. The number of vertices $|\mathcal{V}(s,a)|$ of each polytope 
$\mathcal{P}(s,a)$ can be exponential in the number of successor states, 
and constraint~(\ref{enc:c4-vertex}) must be instantiated for every vertex. 
As a result, the MILP may contain exponentially many constraints in the size 
of the IMDP, making this encoding expensive when state--action pairs have many successors
and motivating the dualization-based approach as follows.

\subsection{Dualization-Based Encoding} \label{sec:dual}

\input{4b_dual}

We now present a compact MILP encoding based on linear programming duality as 
shown in \encref{enc:dual}. The objective~(\ref{enc:obj-dual}) and 
constraints~(\ref{enc:c1-dual})–(\ref{enc:c3-dual}) are identical to those in 
\encref{enc:vertex}.

Instead of enumerating all vertices of the uncertainty polytopes, we encode the 
worst-case choice of transition probabilities directly using dual variables, 
a standard approach in robust optimization~\cite{ben2009robust}.
Here, $\hat{u}_{s,a,s'},\check{u}_{s,a,s'} \ge 0$ are dual variables associated 
with the upper and lower bounds $\hat{P}(s,a,s')$ and $\check{P}(s,a,s')$, 
$\lambda_{s,a}$ is a free dual variable, and $\eta_{s,a}$ is an auxiliary 
variable used to linearize the product $\lambda_{s,a} y_{s,a}$ with the big-$M$ 
method~\cite{mccormick1976computability}.

Constraint~(\ref{enc:c4-dual}) enforces \emph{dual feasibility}: for each 
state–action pair $(s,a)$ and successor $s'$, the dual variables satisfy 
$\lambda_{s,a} - \hat{u}_{s,a,s'} + \check{u}_{s,a,s'} \le x_{s'}$. 
Constraint~(\ref{enc:c5-dual}) is the core robust inequality, bounding the 
value $x_s$ by the dual objective. Intuitively, it ensures that the worst-case 
distribution consistent with the intervals is captured without explicit vertex 
enumeration. 
Finally, constraints~(\ref{enc:c6a-dual})–(\ref{enc:c6b-dual}) implement the 
big-$M$ linearization, guaranteeing $\eta_{s,a}=0$ when $y_{s,a}=0$ and 
$\eta_{s,a}=\lambda_{s,a}$ when $y_{s,a}=1$.

\begin{examp}\label{eg:dual}
We revisit the IMDP and property from \egref{eg:vertex}, now using \encref{enc:dual}. 
For the fast action at $s_0$, constraint~(\ref{enc:c4-dual}) 
introduces dual feasibility inequalities linking 
$\lambda_{s_0,f}, \hat{u}_{s_0,f,s'}, \check{u}_{s_0,f,s'}$ 
to the successor values $x_{s'}$. 
Constraint~(\ref{enc:c5-dual}) then yields the robust bound
\begin{align*}
x_{s_0} \le\; & M(1-y_{s_0,f}) + \eta_{s_0,f} \\
&+ (0.22-\varepsilon)\,\check{u}_{s_0,f,s_2} 
   - (0.22+\varepsilon)\,\hat{u}_{s_0,f,s_2} \\
&+ (0.78-\varepsilon)\,\check{u}_{s_0,f,s_3} 
   - (0.78+\varepsilon)\,\hat{u}_{s_0,f,s_3}.
\end{align*}
Here the auxiliary variable $\eta_{s_0,f}$ linearizes the product 
$\lambda_{s_0,f} y_{s_0,f}$, ensuring that $\lambda_{s_0,f}$ contributes 
only when $y_{s_0,f}=1$.
Solving this encoding yields the same optimal permissiveness as the vertex-enumeration encoding for this example.
\end{examp}

\startpara{Correctness}
The correctness of the MILP in \encref{enc:dual} is stated in the following 
theorem, with the proof provided in the appendix.

\begin{theorem} \label{thm:dual}
Let $\mathcal{M}$ be an IMDP and 
$\varphi=\mathsf{P}_{\ge p}(\Diamond T)$ a property.
An optimal solution to the MILP in \encref{enc:dual} induces a robust,
optimally permissive multi-strategy
$\theta(s)=\{a\in\alpha(s)\mid y_{s,a}=1\}$ for $\varphi$.
Conversely, every robust, optimally permissive multi-strategy can be encoded
by an optimal solution whose objective value equals
$\beta(\theta)=\sum_{s\in S}\sum_{a\in\alpha(s)}y_{s,a}$.
\end{theorem}

\startpara{Adapting \encref{enc:dual} for Other Specifications}
For $\mathsf{P}_{\le p}(\Diamond T)$, robust satisfaction requires an upper
bound on the maximum reachability probability over admissible transition
distributions. Thus, constraint~(\ref{enc:c2-dual}) is replaced by
$x_{s_\iota}\le p$.
Constraint~(\ref{enc:c4-dual}) is replaced by
$\lambda_{s,a} + \hat{u}_{s,a,s'} - \check{u}_{s,a,s'} \ge x_{s'}$. 
Constraint~(\ref{enc:c5-dual}) is replaced by
\(
x_s \ge -M(1-y_{s,a})+\eta_{s,a}
+\sum_{s'\in S}
\bigl(
\hat{P}(s,a,s')\hat{u}_{s,a,s'}
-\check{P}(s,a,s')\check{u}_{s,a,s'}
\bigr).
\)
These constraints correspond to the dual of the maximization over
$\mathcal{P}(s,a)$.

For reward properties $\mathsf{R}_{\bowtie b}(\Diamond T)$, the variables
$x_s$ represent reward-to-go, constraint~(\ref{enc:c2-dual}) is replaced by
$x_{s_\iota}\bowtie b$, and constraint~(\ref{enc:c3-dual}) is set to
$x_s=0$ for all $s\in T$. The immediate reward $r(s,a)$ is added to the
corresponding robust inequality. The minimum-dual formulation is used for
lower-bound properties and the maximum-dual formulation for upper-bound
properties.
\startpara{Scalability}
The dualization-based encoding may introduce more constraints in small 
instances such as the IMDP in \figref{fig:example}. 
The relative computational benefit depends not only on the overall size of the 
IMDP, but also on its local branching structure. When each state--action pair 
has few successors, explicit vertex enumeration may remain compact and can 
outperform the dual formulation. As the number of successors grows, however, 
the number of vertices can increase exponentially, whereas the dual formulation 
grows only linearly in the number of successor transitions. Its advantage 
therefore becomes most pronounced for models with larger branching factors, 
as demonstrated in \sectref{sec:exp}.

%% file: 4a_vertex.tex
\begin{encoding}{Vertex Enumeration for $\mathsf{P}_{\ge p}(\Diamond T)$}\label{enc:vertex}
\begingroup
\setlength{\abovedisplayskip}{2pt}
\setlength{\belowdisplayskip}{2pt}
\begin{subequations}
\begin{align}
\maximize \sum_{s\in S}\sum_{a\in \alpha(s)} y_{s,a}
\label{enc:obj-vertex} \\
\text{subject to:} \notag \\[2pt]
\forall s\in S:\ \sum_{a\in \alpha(s)} y_{s,a} \ge 1
\label{enc:c1-vertex} \\
x_{s_\iota} \ge p
\label{enc:c2-vertex} \\
\forall s \in T:\ x_s = 1
\label{enc:c3-vertex} \\
\forall s\in S\setminus T,\ \forall a\in \alpha(s),\ 
\forall v\in \mathcal{V}(s,a): \nonumber \\
x_s \le \sum_{s'\in S} P^{(v)}(s,a,s') \cdot x_{s'} + M(1-y_{s,a})
\label{enc:c4-vertex}
\end{align}
\end{subequations}
\endgroup
\end{encoding}

%% file: 4b_dual.tex
\begin{encoding}{Dualization for $\mathsf{P}_{\ge p}(\Diamond T)$}\label{enc:dual}
\begingroup
\setlength{\abovedisplayskip}{2pt}
\setlength{\belowdisplayskip}{2pt}
\begin{subequations}
\begin{align}
\maximize \ \sum_{s\in S}\sum_{a\in \alpha(s)} y_{s,a}
\label{enc:obj-dual} \\
\text{subject to:} \notag \\[2pt]
\forall s\in S:\ \sum_{a\in \alpha(s)} y_{s,a} \ge 1
\label{enc:c1-dual} \\
x_{s_\iota} \ge p
\label{enc:c2-dual} \\
\forall s \in T:\ x_s = 1
\label{enc:c3-dual} \\
\forall (s,a,s'):\ \lambda_{s,a} - \hat{u}_{s,a,s'} + \check{u}_{s,a,s'} \le x_{s'}
\label{enc:c4-dual} \\
\forall (s,a):\ x_s \le M(1-y_{s,a}) + \eta_{s,a} \nonumber \\
+ \sum_{s'\in S} \bigl(\check{P}(s,a,s')\,\check{u}_{s,a,s'} 
- \hat{P}(s,a,s')\,\hat{u}_{s,a,s'} \bigr) 
\label{enc:c5-dual} \\
\forall (s,a):\ -M\,y_{s,a} \le \eta_{s,a} \le M\,y_{s,a}
\label{enc:c6a-dual} \\
\forall (s,a):\ \lambda_{s,a} - M(1-y_{s,a}) \le \eta_{s,a} \nonumber \\
\le \lambda_{s,a} + M(1-y_{s,a})
\label{enc:c6b-dual}
\end{align}
\end{subequations}
\endgroup
\end{encoding}

%% file: 5_exp.tex
We implemented the proposed synthesis methods using PRISM~\cite{kwiatkowska2011prism} for IMDP parsing and Gurobi~\cite{gurobi} for solving the resulting MILPs. As a baseline for comparison, we used PRISM's default robust value iteration method for single-policy synthesis on IMDPs. For each benchmark, we used the same IMDP and specification for the two MILP encodings and the single-policy baseline.%
\footnote{Benchmark models and MILP encodings are available at 
\url{https://www.prismmodelchecker.org/files/lcss-imdp/}}.  
Experiments were conducted on a 2.8~GHz Quad-Core Intel Core~i7 laptop with 16~GB RAM.

We evaluate four IMDP benchmark domains: aircraft collision avoidance (ACA)~\cite{schnitzer2025certifiably}, semi-autonomous vehicle navigation (SAV)~\cite{junges2016safety}, obstacle navigation (OBS) inspired by FrozenLake~\cite{brockman2016openai}, and warehouse navigation (WH)~\cite{chen2022multi}. ACA, SAV, and OBS use probabilistic reachability specifications $P_{\geq p}(\Diamond T)$, while WH uses an expected-reward specification $R_{\leq b}(\Diamond T)$. The benchmark parameters respectively control branching factor, grid size, grid/movement granularity, and checkpoint traversal length.

\input{5a_table}

\tabref{tab:results} reports IMDP size, baseline runtime and permissiveness, and the MILP size and runtime of both encodings.
We report \emph{normalized choice-state permissiveness}, defined as
\(
\frac{\sum_{s\in S_c} |\theta(s)|}
     {\sum_{s\in S_c} |\alpha(s)|}
\)
where $S_c$ denotes the set of choice states, i.e., reachable non-target states with at least two available actions.
A deterministic single policy enables exactly one action at each choice state and thus provides a natural low-permissiveness reference.

\emph{Permissive synthesis retains substantially more action choices.}
Across all benchmark instances, the synthesized multi-strategies achieve substantially higher permissiveness than the single-policy baseline while satisfying the same robust specification. Both MILP encodings attain the same normalized choice-state permissiveness 
on every instance.

\emph{The relative efficiency of the two encodings depends on branching factor.}
Vertex enumeration is competitive when each state--action pair has few successors, whereas its number of constraints grows rapidly as branching increases. This is most evident in ACA, where dualization produces substantially smaller MILPs and becomes faster for the larger instances. For SAV, OBS, and WH, where branching is smaller, the additional real variables and constraints introduced by dualization can outweigh this benefit, making vertex enumeration faster.

\emph{Permissive synthesis can remain computationally competitive with single-policy synthesis.}
Although the single-policy baseline is inexpensive on several small instances, its runtime becomes substantial on some larger models. In several cases of OBS and WH, the MILP formulations are even faster while providing substantially greater flexibility.

%% file: 5a_table.tex
\begin{table*}[t]
\centering
\resizebox{\textwidth}{!}{%
\begin{tabular}{@{}cc|rr|rr|rrrrr|rrr@{}}
\toprule
\multicolumn{2}{c}{Domain} 
& \multicolumn{2}{c}{IMDP Size} 
& \multicolumn{2}{c}{Single Policy}
& \multicolumn{5}{c}{Vertex Enumeration (\encref{enc:vertex})}
& \multicolumn{3}{c}{Dualization (\encref{enc:dual})} \\
\cmidrule(lr){1-2}
\cmidrule(lr){3-4}
\cmidrule(lr){5-6}
\cmidrule(lr){7-11}
\cmidrule(lr){12-14}
Name & Parameters 
& \#States & \#Trans 
& Time(s) & Perm(\%)
& Perm(\%) & \#Binary & \#Real & \#Constraints & Time(s)
& \#Real & \#Constraints & Time(s) \\
\midrule

\multirow{3}{*}{ACA}
& 10 & 328 & 4,739
& 0.03 & 25.11
& 98.40 & 984 & 390 & 110,398 & 4.3
& 11,774 & 31,010 & 7.7 \\

& 12 & 453 & 6,498
& 0.04 & 26.25
& 98.72 & 1,203 & 529 & 585,749 & 22.7
& 15,855 & 41,507 & 5.6 \\

& 14 & 640 & 9,583
& 0.04 & 26.19
& 99.33 & 1,632 & 733 & 2,814,507 & 145.4
& 23,070 & 60,163 & 4.3 \\
\midrule

\multirow{3}{*}{SAV}
& 5 & 279 & 999
& 0.03 & 33.33
& 96.73 & 727 & 363 & 1,370 & 0.6
& 3,731 & 8,274 & 5.3 \\

& 6 & 411 & 1,503
& 0.04 & 33.33
& 97.48 & 1,099 & 539 & 2,050 & 46.3
& 5,615 & 15,755 & 218.6 \\

& 7 & 567 & 2,103
& 0.04 & 33.33
& 97.47 & 1,543 & 747 & 2,858 & 2,564.7
& 7,859 & 22,071 & 4,030.2 \\
\midrule

\multirow{3}{*}{OBS}
& 5; 1000 & 73,025 & 73,299
& 154.1 & 23.68
& 67.11 & 73,083 & 73,030 & 146,471 & 2.1
& 365,789 & 951,108 & 4.8 \\

& 6; 100 & 11,336 & 11,760
& 2.6 & 24.14
& 74.14 & 11,424 & 11,342 & 23,324 & 1.4
& 57,704 & 115,840 & 3.7 \\

& 6; 1000 & 113,036 & 113,460
& 468.0 & 24.14
& 74.14 & 113,124 & 113,042 & 226,724 & 3.2
& 566,204 & 1,132,840 & 14.8 \\
\midrule

\multirow{3}{*}{WH}
& 250 & 2,762 & 3,785
& 0.8 & 50.00
& 99.95 & 3,768 & 2,762 & 6,549 & 0.2
& 17,868 & 36,861 & 1.1 \\

& 500 & 5,512 & 7,535
& 2.6 & 50.00
& 99.98 & 7,518 & 5,512 & 13,049 & 0.4
& 35,618 & 63,261 & 1.8 \\

& 1000 & 11,012 & 15,035
& 10.2 & 50.00
& 99.98 & 15,018 & 11,012 & 26,049 & 0.7
& 71,118 & 146,261 & 3.0 \\

\bottomrule
\end{tabular}%
}
\caption{Experimental results for single-policy synthesis and permissive synthesis via two encodings. Binary-variable counts and permissiveness are identical for both encodings and are shown once.}
\label{tab:results}
\end{table*}

%% file: 6_conclu.tex
We presented a framework for robust permissive controller synthesis on IMDPs, 
enabling controllers that guarantee specification satisfaction under all 
admissible transitions while preserving multiple action choices for runtime 
flexibility. The problem was formulated as mixed-integer linear programs with 
two encodings: a conceptually simple vertex-enumeration method and a 
dualization-based method whose size grows linearly with the number of successor 
transitions. Experiments on four benchmark domains showed that both encodings 
synthesize optimally permissive robust multi-strategies, while their relative 
computational efficiency depends on the local branching structure of the IMDP.

Our formulation considers deterministic memoryless compliant strategies and 
optimizes permissiveness by equally weighting enabled state--action pairs. 
Alternative strategy classes and application-dependent notions of permissiveness 
are natural directions for extending the framework.

Future work includes evaluating the framework on a broader range of IMDP applications 
and integrating permissive synthesis with learning-based control. In particular, 
robust permissive strategies align naturally with shielding in safe reinforcement 
learning, suggesting opportunities for safety-preserving integration with 
learning-based agents that must adapt across tasks and environments.

%% file: 7_appendix.tex
\startpara{Correctness of \encref{enc:vertex}}
We first establish two auxiliary lemmas and then prove \thmref{thm:vertex}.
We assume standard preprocessing removes non-target end components.

\begin{lemma} \label{lem:vertex}
For any $(s,a)$ and vector $x\in\mathbb{R}^S$, the extremum of 
$\sum_{s'} P(s,a,s')\,x_{s'}$ over $P\in\mathcal{P}(s,a)$ is attained 
at some $P^{(v)}(s,a,\cdot)\in\mathcal{V}(s,a)$.
\end{lemma}

\begin{proof}
For fixed $(s,a)$ and $x$, the expression 
$\sum_{s'} P(s,a,s')\,x_{s'}$ is linear and 
$\mathcal{P}(s,a)$ is a convex polytope. Hence any extremum is attained 
at a vertex $P^{(v)}(s,a,\cdot)\in\mathcal{V}(s,a)$.
\end{proof}

\begin{lemma}\label{lem:ineq}
Let $\theta$ be a multi-strategy and let $\bar{x}$ denote its robust 
reachability values. For $s\notin T$,
\[
\bar{x}_s =
\min_{a\in\theta(s)}
\min_{P\in\mathcal{P}(s,a)}
\sum_{s'\in S}P(s,a,s')\,\bar{x}_{s'},
\]
with $\bar{x}_s=1$ for $s\in T$. Moreover, any vector $x$ satisfying
\[
x_s \le
\min_{a\in\theta(s)}
\min_{P\in\mathcal{P}(s,a)}
\sum_{s'\in S}P(s,a,s')\,x_{s'}
\]
for all $s\notin T$, together with $x_s=1$ on $T$, satisfies 
$x\le\bar{x}$.
\end{lemma}

\begin{proof}
After preprocessing, the induced robust MDP has no non-target end component. 
The first statement follows from the standard Bellman equations for minimum 
reachability, and the second from monotonicity of the Bellman operator
~\cite{puterman1994markov}.
\end{proof}

\emph{Proof of \thmref{thm:vertex}:}
(Soundness) Let $(x,y)$ be an optimal solution to \encref{enc:vertex} and define
$\theta(s)=\{a\in\alpha(s)\mid y_{s,a}=1\}$.
Constraints~(\ref{enc:c1-vertex})--(\ref{enc:c3-vertex}) enforce non-deadlock,
$x_{s_\iota}\ge p$, and $x_s=1$ for $s\in T$.
For every allowed action, constraint~(\ref{enc:c4-vertex}) holds at all vertices
of $\mathcal{P}(s,a)$. By \lemref{lem:vertex}, this is equivalent to the robust
Bellman inequality, and \lemref{lem:ineq} therefore gives $x\le\bar{x}$.
Hence $\bar{x}_{s_\iota}\ge x_{s_\iota}\ge p$, so
$(\mathcal{M},\theta)\models_{\mathrm{rob}}\varphi$.
Since the objective equals $\beta(\theta)$, optimality yields an optimally
permissive multi-strategy.

(Completeness) Conversely, let $\theta$ be robust and optimally permissive,
let $y$ encode $\theta$, and set $x=\bar{x}$.
The Bellman equations and \lemref{lem:vertex} imply
constraint~(\ref{enc:c4-vertex}), while robust satisfaction and the definition
of $\theta$ give the remaining constraints.
Thus $(x,y)$ is feasible with objective value $\beta(\theta)$.
Since $\theta$ is optimally permissive, the assignment is optimal.
{\hfill\scalebox{0.85}{$\blacksquare$}\vskip4pt}

\startpara{Correctness of \encref{enc:dual}}
We first establish an auxiliary lemma and then prove \thmref{thm:dual}.

\begin{lemma}\label{lem:dual-eq}
Fix $(s,a)$ and $x\in\mathbb{R}^S$. The robust inequality
\[
x_s \le
\min_{P\in\mathcal{P}(s,a)}
\sum_{s'\in S}P(s,a,s')x_{s'}
\]
holds iff there exist
$\hat{u}_{s,a,s'},\check{u}_{s,a,s'}\ge0$ and
$\lambda_{s,a}\in\mathbb{R}$ such that
\[
\lambda_{s,a}-\hat{u}_{s,a,s'}+\check{u}_{s,a,s'}
\le x_{s'} \qquad (\forall s'\in S)
\]
and
\[
x_s \le \lambda_{s,a}
+\sum_{s'\in S}
\bigl(
\check{P}(s,a,s')\check{u}_{s,a,s'}
-\hat{P}(s,a,s')\hat{u}_{s,a,s'}
\bigr).
\]
\end{lemma}

\begin{proof}
The inner minimization is a linear program over interval and simplex
constraints. Its dual has nonnegative variables $\hat{u},\check{u}$ and a
free variable $\lambda$, yielding the displayed conditions. The result
follows from strong LP duality~\cite{ben2009robust}.
\end{proof}

\emph{Proof of \thmref{thm:dual}:}
(Soundness) Let $(x,y,\hat{u},\check{u},\lambda,\eta)$ be an optimal solution
to \encref{enc:dual} and define
$\theta(s)=\{a\in\alpha(s)\mid y_{s,a}=1\}$.
Constraints~(\ref{enc:c1-dual})--(\ref{enc:c3-dual}) enforce non-deadlock,
$x_{s_\iota}\ge p$, and $x_s=1$ on $T$.
By \lemref{lem:dual-eq},
constraints~(\ref{enc:c4-dual})--(\ref{enc:c5-dual}) enforce the robust
Bellman inequality for each allowed action, while
(\ref{enc:c6a-dual})--(\ref{enc:c6b-dual}) implement the big-$M$
linearization.
Thus \lemref{lem:ineq} gives $x\le\bar{x}$, so
$\bar{x}_{s_\iota}\ge p$ and
$(\mathcal{M},\theta)\models_{\mathrm{rob}}\varphi$.
Optimality of the objective yields optimal permissiveness.

(Completeness) Conversely, let $\theta$ be robust and optimally permissive,
let $y$ encode $\theta$, and set $x=\bar{x}$.
For each allowed $(s,a)$, \lemref{lem:dual-eq} provides dual variables
satisfying constraints~(\ref{enc:c4-dual})--(\ref{enc:c5-dual}).
Choose $\eta_{s,a}=\lambda_{s,a}$ when $y_{s,a}=1$ and
$\eta_{s,a}=0$ otherwise, so the big-$M$ constraints hold.
The resulting assignment is feasible with objective value $\beta(\theta)$
and is therefore optimal.
{\hfill\scalebox{0.85}{$\blacksquare$}\vskip4pt}